\documentclass{article}

\usepackage[preprint]{neurips_2022}

\usepackage[utf8]{inputenc} % allow utf-8 input
\usepackage[T1]{fontenc}    % use 8-bit T1 fonts
\usepackage{hyperref}       % hyperlinks
\usepackage{url}            % simple URL typesetting
\usepackage{booktabs}       % professional-quality tables
\usepackage{amsfonts}       % blackboard math symbols
\usepackage{nicefrac}       % compact symbols for 1/2, etc.
\usepackage{microtype}      % microtypography
\usepackage{xcolor}         % colors
\hypersetup{pdfborder=0 0 0}
\usepackage{bm}
\usepackage{amsmath}
\usepackage{graphicx}
\usepackage{lscape}
\usepackage{float}
\usepackage{longtable}
\usepackage{adjustbox}

\newcommand{\reticl}{RetICL}

\title{In-context Learning with Retrieved Demonstrations for Language Models: A Survey}

\author{
    Man Luo$^{1}$ \quad Xin Xu$^2$ \quad \textbf{Yue Liu}$^2$ \quad \textbf{Panupong Pasupat}$^2$ \quad \textbf{Mehran Kazemi}$^2$\\
    \textsuperscript{1} Arizona State University \quad \textsuperscript{2} Google Research\\
    \texttt{mluo26@asu.edu} \\
    \texttt{\{xxujasmine, yliujune,
ppasupat, mehrankazemi\}@google.com}
}

\begin{document}

\maketitle

\begin{abstract}

Language models, especially pre-trained large language models, have showcased remarkable abilities as few-shot in-context learners (ICL), adept at adapting to new tasks with just a few demonstrations in the input context. 
However, the model's ability to perform ICL is sensitive to the choice of the few-shot demonstrations.
Instead of using a fixed set of demonstrations, one recent development is to \emph{retrieve} demonstrations tailored to each input query.
The implementation of demonstration retrieval is relatively straightforward, leveraging existing databases and retrieval systems. This not only improves the efficiency and scalability of the learning process but also has been shown to reduce biases inherent in manual example selection. In light of the encouraging results and growing research in ICL with retrieved demonstrations, we conduct an extensive review of studies in this area. In this survey, we discuss and compare different design choices for retrieval models, retrieval training procedures, and inference algorithms.
 
\end{abstract}

% \tableofcontents

% \newpage

\section{Introduction}

Few-shot in-context learning (ICL) is the ability of large language models (LLMs) to perform a new task when a few input-output examples, or \emph{demonstrations}, for the new task are given alongside the actual task input.
Importantly, the model parameters do not have to be fine-tuned towards the new task.
% ICL is an emergent behavior of pre-trained large language models (LLMs), meaning that the ability is gained when the model is scaled up to certain model and data sizes, without an explicit training objective to perform ICL~\citep{brown2020language}.\footnote{Note that language models, including smaller ones, can be explicitly trained to perform ICL~\citep{min2022metaicl}.}
ICL is popularized by the work on pre-trained large language models, which can perform ICL without being trained to do so~\citep{brown2020language},
though smaller language models can also be explicitly trained to perform ICL~\citep{min2022metaicl}.

ICL presents several advantages over the conventional methodology for adapting language models to a downstream task, which typically involves initial pre-training followed by subsequent fine-tuning. One significant merit of ICL is the circumvention of fine-tuning, which might not always be possible due to limited access to the model parameters or constraints on computational resources~\citep{brown2020language}. Furthermore, ICL avoids common issues associated with fine-tuning, such as overfitting\citep{ying2019overview,kazemi2023understanding}. Compared to parameter-efficient fine-tuning methods (PEFT)~\citep{hu2021lora,dettmers2023qlora,lester2021power}, ICL is computationally cheaper and remain the model parameters unchanged thus preserving the generality of the LLMs. 

Early ICL implementations use a fixed set of demonstrations for each target task. These demonstrations could be hand-crafted by human~\citep{hendrycks2021measuring,wei2022chain,kazemi2023lambada}, randomly chosen from training data~\citep{brown2020language,lewkowycz2022solving}. 
Beyond random selection, there are more advanced selection processes based on metrics such as complexity~\citep{fu2022complexity}, diversity~\citep{li2023finding}, difficulty~\citep{drozdov2023parade}, concept learning~\citep{wang2023large} and perplexity~\citep{gonen2023demystifying}. 
Importantly, the demonstrations remain context-insensitive (i.e. the same demonstrations are used regardless of the query) which could hinder unlocking the true potential of the LLMs. 
The effectiveness of such demonstrations is influenced by factors such as the quality, quantity, and ordering of the demonstrations.

\begin{figure*} 
    \centering
    \includegraphics[width=\linewidth]{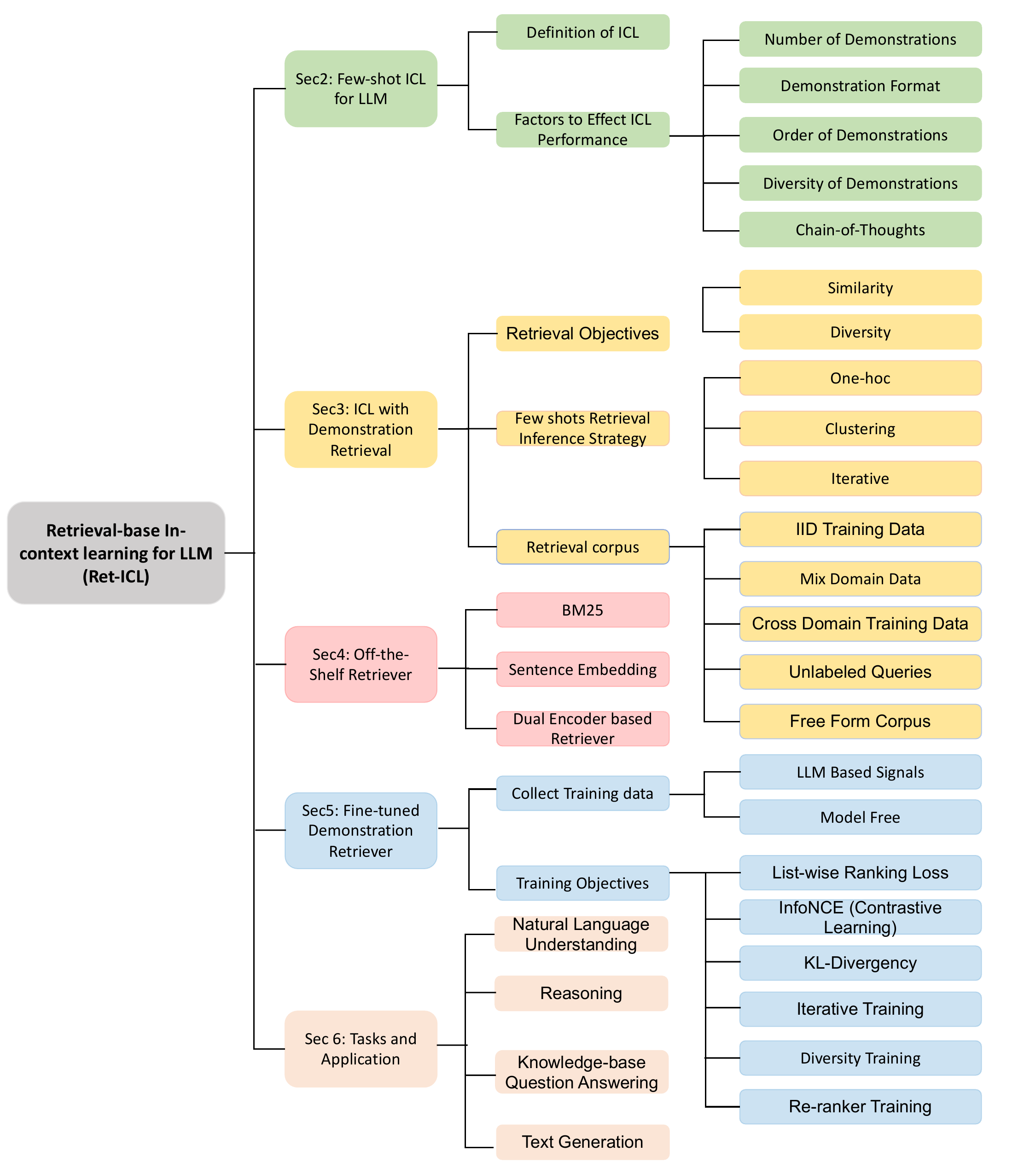}
    \caption{Structure of the Survey.}
    \label{fig:ret-survey}
\end{figure*}

Retrieval-based ICL (\reticl) presents a paradigm shift in the optimization of language model performance, moving beyond static, pre-defined demonstration sets to a dynamic, context-sensitive approach. At the heart of this innovation is the concept of \emph{adaptive demonstration selection}, where a specialized retriever intelligently curates tailored demonstrations for each specific task input. This method has not only consistently outshined approaches relying on random or static hand-crafted demonstrations but has also demonstrated a remarkable resilience to a variety of influencing factors.

The efficacy of \reticl\ pivots on the ``relevance'' and ``usefulness'' of the demonstrations it selects, a process intricately influenced by multiple elements. These include the nature of the retriever—ranging from general off-the-shelf models to finely-tuned, domain-specific variants—the source and diversity of the retrieval corpus, the retriever's objectives (focusing on either similarity or diversity), and the strategies for integrating multiple demonstrations.
Over the past two years, numerous and sometimes concurrent works have studied \reticl\ each with different terminology and with variations in problem definition and subsequent methodologies, making it difficult to comprehend the current state of research and practice in \reticl, especially for newcomers to the field. In this comprehensive survey, we meticulously analyze 22 seminal papers in the field of \reticl, as detailed in Table~\ref{tab:related_work}, and provide a categorization of their main building blocks (See Figure \ref{fig:ret-survey}). Our work not only provides a thorough synthesis of existing research but also underscores the areas where \reticl\ significantly surpasses previous ICL methods, and illuminates many paths forward for future innovations in this area, thus serving as a critical resource for ICL.

\begin{table}
    \centering
    \resizebox{0.99\linewidth}{!}{
    \setlength{\tabcolsep}{5pt}
    \begin{tabular}{@{}p{0.1\linewidth}p{0.1\linewidth}p{0.2\linewidth}p{0.2\linewidth}p{0.2\linewidth}p{0.2\linewidth}p{0.1\linewidth}@{}}
        \toprule
        \textbf{Paper} & \textbf{LLMs}  & \textbf{Retrieval Method}  & \textbf{Retrieval Corpus} & \textbf{Evaluation Tasks}& \textbf{ Retriever Training} & \textbf{Retrieval Strategy}\\
        \toprule
    SYNCHRO
    MESH~\citeyear{poesia2021synchromesh} & GPT-3 & 
SBERT + Target Similarity & In Domain & CodeGen & Target Similarity Tuning & One-hoc \\
    \midrule 
    KATE~\citeyear{liu2022makes} & GPT-3 & RoBERTa+kNN & In Domain & SA, Table2Text, QA & Sentence Similarity Embedding & One-hoc \\
    \midrule
    EPR~\citeyear{rubin2021learning} & GPT-J, CODEX, GTP-3 & SBERT, BM25, Fine-tuned Retriever & In Domain & SP& InfoNCE Loss(contrastive learning)& One-hoc \\
    \midrule
    Z-ICL~\citeyear{lyu2022z} & GPT-J, GPT-NeoX, GPT-3 & SimCSE (Sentence Embeddings) & Free Form Corpus & SA & Sentence Embeddings Similarity & One-hoc \\
    \midrule
    Mem
    Prompt~\citeyear{madaan2022memory}& GPT-3 & Query Transformation + SBERT & In Domain with Human Feedback & Word Reasoning, Ethical Reasoning & Sentence Similarity Embeddings  & One-hoc\\
    \midrule
    Teach
    Me~\citeyear{dalvi2022towards} & T5-11B & BM25 &In Domain with Human Feedback &  QA & Term-based Similarity & One-hoc\\
    \midrule
    IC-DST~\citeyear{hu2022context} & GPT-3 & Fine-tuned SBERT Retriever & In Domain & DST-to-SQL & Contrastive Learning & One-hoc \\
    \midrule
    Vote-k~\citeyear{hongjin2022selective} & GPT-Neo, GPT-J,GPT-3, CODEX & SBERT & Selected subsets of In Domain, labeled &SA, NLI, QA, Summ, CSR, RC &  Sentence Embeddings Similarity & One-hoc \\
    \midrule 
    Auto-CoT~\citeyear{zhang2022automatic} & GPT-3 & Clustering with SBERT & LLM Generated CoT for In Domain & MathR, QA & Sentence Embeddings Similarity & Clustering   \\
    \midrule
    XRICL~\citeyear{shi2022xricl} & CODEX & Fine-tuned Retriever Combined with a Re-ranker & Cross Domain & Text-to-SQL & Distillation by KL Divergence & One-hoc \\
    \midrule 
    DSP~\citeyear{khattab2022demonstrate} & GPT-3 & ColBERTv2 & In Domain & QA & Token Embeddings Similarity & Iterative \\
    \midrule
    PARC~\citeyear{Nie2022CrossLingualRA} & mBERT & Multilingual SBERT & Cross Domain 
    & SA, TopicC, NLI & Sentence Embeddings Similarity  & One-hoc \\
    \midrule
    Cover-LS~\citeyear{levy2022diverse} & CODEX, T5-large & BM25 and Diversity Selection &  In Domain & SP & Diversity Training & Iterative \\
    \midrule 
    Prompt
    PG~\citeyear{lu2022dynamic} & GPT-3 & Fine-tuned Retriever &  In Domain &  MathR  & Policy Gradient  & One-hoc \\
    \midrule 
    Dynamic Least-to-Most~\citeyear{drozdov2022compositional} & DODEX & Tree Structure Similarity & In Domain & SP & Diversity and Lexical Overalp & Iterative\\
    \midrule
    Self-
    Prompting~\citeyear{li2022self} & Instruct
    GPT, CODEX & SBERT with Clustering & Free Form Corpus & QA & Sentence Embeddings Similarity & Clustering\\
    \midrule
    CEIL~\citeyear{ye2023compositional}& GPT-NEO, GPT2-XL, CODEX & BM25, BERT, Fine-tuned Retriever & In Domain & SA, PD, NLI, CSR, QA, CodeGen, and SP & Diversity Training & Iterative \\
    \midrule
    R-BM25~\citeyear{agrawal2023context} & XGLM-7.5G & BM25, BM25 and Term Recall based rerank In Domain & In Domain & MT & Term-based Similarity & One-hoc \\
    \midrule
    ICL-MC~\citeyear{sia2023context} & GPTNeo-2.7B, XGLM-2.9G, BLOOM-3B & BM25 & In Domain Cross Domain & MT & Term-based Similarity & One-hoc\\
    \midrule
    ICL-ML~\citeyear{milios2023context} & OPT-13/175B, LLaMA-7/70B & SBERT & In Domain & Intent Classification & Sentence Embeddings Similarity & One-hoc \\
    \midrule
    UP
    RISE~\citeyear{cheng2023uprise} & GPT-Neo, BLOOM, OPT, GPT-3 & Fine-tuned Retriever & Cross Domains Human Feedback& RC, QA, NLI, SA, CSR, CR, PD & Contrastive Learning & One-hoc \\
    
    \bottomrule
    \end{tabular}
    }
\end{table}

% {\footnotesize
% \begin{longtable}{@{}p{0.1\linewidth}p{0.1\linewidth}p{0.2\linewidth}p{0.2\linewidth}p{0.2\linewidth}p{0.2\linewidth}p{0.1\linewidth}@{}} 
\begin{table}
    \centering
    \resizebox{0.99\linewidth}{!}{
    \setlength{\tabcolsep}{5pt}
    \begin{tabular}{@{}p{0.1\linewidth}p{0.1\linewidth}p{0.2\linewidth}p{0.2\linewidth}p{0.2\linewidth}p{0.2\linewidth}p{0.1\linewidth}@{}}
        \toprule
        \textbf{Paper} & \textbf{LLMs}  & \textbf{Retrieval Method}  & \textbf{Retrieval Corpus} & \textbf{Evaluation Tasks}& \textbf{ Retriever Training} & \textbf{Retrieval Strategy}\\
        \toprule
    Dr.ICL~\citeyear{luo2023dr} & PaLM, Flan-PaLM & BM25, GTR, Fine-tuned Retriever &  In Domain & QA, NLI, MathR & Contrastive Learning & One-hoc \\
    \midrule
    LLM-R~\citeyear{wang2023learning} & LLaMA-7B & Fine-tuned Retriever and Reward Model  & Mix Domain  & QA, CSR, CR, PD, RC, SA, D2T, Summ, NLI & Contrastive learning + KL & One-hoc \\
    \midrule
    UDR~\citeyear{li2023unified} & GPT-J, GPT-Neo, CODEX, GPT-3 & Fine-tuned Unified Retriever with Iterative Data Mining & Mix Domain & SA, TC, CSR, NLI, SP, StoryGen, Summ, D2T  & Contrastive Learning & One-hoc \\
    \midrule
    MoT~\citeyear{li2023mot} & ChatGPT & SBERT with Clustering and Filtering, LLM as Retriever & Unlabelled Queries with LLMs CoT & MathR, NLI, CSR, QA  & Sentence Embeddings Similarity & Clustering \\
    \midrule 
    RetICL~\citeyear{scarlatos2023reticl} & ~CODEX & Iterative LSTM with Reinforcement Learning & In Domain  & MathR & PPO and GAE & Iterative \\
    \midrule
    Ambig-ICL~\citeyear{gao2023ambiguity} & Flan-PaLM & Fine-tuned multilingual T5, and a Filtering Algorithm Based on LLM feedback & In Domain & TC, SA & Contrastive Learning & One-hoc\\
    \bottomrule
    \end{tabular}
    }
    \caption{Comparison with Related Work. Abbreviation for Evaluation Tasks: CodeGen (code generation), SA (sentiment analysis), Table2Text (Table to Text generation), QA (question answering), SP (semantic parsing),  DST (Dialogue State Tracking), D2T (Data-to-Text),  Summ ( Summarization), CSR ( commonsense reasoning),  RC (reading comprehension), NLI (natural language inference),  CR (Coreference Resolution), MathR (mathematical reasoning), PD (paraphrase detection),  TQA (Table Question Answering), TC (Topic Classification), StoryGen (Story Generation), MT (Machine Translation)
    }
    \label{tab:related_work} 
\end{table}

% \subsection{Annoation} 

% \begin{itemize}
% \item (L)LM: (Large) Language Models.
% \item ICL: In-context learning: few demonstrations and a query are given to LM and LM answer the query by learning from the demonstration. 
% \item Retriever: A system to retrieve demonstrations. 
% \item Reranker: A model to rerank the demonstrations given by a retriever. 
% \item Query (Q): The inference query that need to be answered by a language model. 
% \item Demonstrations (D): The in-context learning examples. 
% \item TD: Training Data: Annotated data for a task with input and output. 
% \end{itemize}

\section{Few-shot In-context Learning for Language Models} \label{sec:few-shot-icl}

Language models (LMs) \citep{zhao2023survey,rosenfeld2000two} are probabilistic models that assign probabilities to sequences of words and are essential components in many tasks. Let $s$ represent a sequence of words (e.g., a sentence) and $w_1, w_2, \dots, w_n$ represent the tokens in the sequence. Based on the chain rule, the probability $p(s)$ can be decomposed into the following product of probabilities:
\[p(s)=p(w_1)p(w_2 \mid w_1)\dots p(w_n \mid w_1, \dots, w_{n-1})=\prod_{k=1}^{n}p(w_k \mid w_{1}, \dots, w_{k-1})\]
where each element in the product corresponds to the probability of a token given the previous tokens. Based on the above decomposition, an LM can be constructed by learning the probability of the next token given the previous ones. 

Earlier LMs were mostly based on N-gram models which are based on the Markovian assumption that the next token only depends on the recent context \citep{jurafsky2021}. Based on this assumption, $p(w_k \mid w_{1}, \dots, w_{k-1})$ is approximated, e.g., by $p(w_k \mid w_{k-2}, w_{k-1})$ in the case of a bi-gram model; $p(w_k \mid w_{k-2}, w_{k-1})$ is then approximated statistically based on the number of times $w_k$ appeared after  $w_{k-2}$ in a large corpora of text, $w_{k-1}$ divided by the total number of times $w_{k-2}, w_{k-1}$ appeared in the corpora. 

With the advent of word embeddings \citep{bengio2000neural,mikolov2013efficient}, neural approaches to language modeling gained more popularity, in which a neural network is used to predict the next token probability. The use of powerful neural networks such as long-short term memory (LSTM) models \citep{hochreiter1997long} and Transformer models \citep{vaswani2017attention} allowed for predicting the next token probability based on a much longer and a variable length context, thus enabling better estimation of $p(w_k \mid w_{1}, \dots, w_{k-1})$.

The increased power of neural LMs led to a new learning paradigm for NLP problems. Historically, the dominant learning paradigm for NLP problems was to train models on task-specific data from scratch. Consequently, for each new task, the model had to learn everything from scratch. This often resulted in poor generalization, especially in the cases where previously unobserved vocabulary was observed at the test time. In the subsequent paradigm, an LM was first pre-trained on a large corpora of text making it learn about how language works and gain a vast amount of knowledge about the world \citep{petroni2019language,lin2020birds,sung2021can,yuan2023tasklama}; the pre-trained LM (PLM) was then further finetuned on data from the new tasks \citep{sarzynska2021detecting,devlin2018bert} thus teaching the general PLM the specifics of the new task. This paradigm often resulted in faster learning and higher predictive performance. It was later shown that further finetuning a PLM on multiple tasks leads to better transfer of knowledge across tasks and may lead to better performance on new tasks \citep{raffel2020exploring}.

\subsection{In-Context Learning}
As the scale of the PLMs and the scale of the datasets on which these models were pre-trained increased -- leading to pre-trained Large Language Models (LLMs), it was discovered that pre-trained LLMs (hereafter, referred to as \emph{LLMs} for brevity) have a remarkable capability of learning in-context from a few demonstrations \citep{brown2020language}. That is, LLMs were shown to be able to adapt to new tasks by only seeing a few examples of the new task in their input, as opposed to needing additional training data or fine-tuning. This is typically referred to as \emph{few-shot in-context learning}. 

Let $\mathcal{T}$ be a task and $q_* \sim \mathcal{T}$ represent a sample query from this task for which we would like to find an answer using an LLM. In the case of few-shot learning, we find or construct multiple demonstrations $\{d_1, \dots, d_k\}$ where each demonstration $d_i = (q_i, a_i)$ contains a query $q_i \sim \mathcal{T}$ and the answer $a_i$ to that query, and feed an input of the form \[q_1~a_1~\dots~q_k~a_k~q_*\]
to the LLM. The input is typically referred to as \emph{prompt}. It is common to add some separator tokens to the prompt so the boundaries of the demonstrations and the questions and answers within those demonstrations are clear. An example prompt will then be as follows:
\[\mathrm{Demonstration~1:~} \mathrm{Query:}~q_1,~~\mathrm{Answer:}~a_1\]
\[...\]
\[\mathrm{Demonstration~k+1:~}\mathrm{Query:}~q_*,~~\mathrm{Answer:}\]

Seeing the demonstrations as a few examples of the task, LLMs learn from the demonstrations in context (without any weight updates) and use a similar pattern to provide an answer to the query $q_*$. Few-shot learning is a remarkable capability of LLMs given that they are not trained on such data during their pre-training. While LLMs show strong few-shot learning capabilities off-the-shelve, it has been shown that warming them up by finetuning them on few-shot data from multiple tasks will further boost their few-shot learning capability \citep{min2022metaicl,chen2022improving,radford2019language}.

Another remarkable ICL capability of LLMs is to learn from in-context instructions: finetuning LLMs on instructions from multiple tasks makes them learn to follow instructions for new tasks \citep{ouyang2022training,longpre2023flan,zhang2023instruction}. In this case, commonly known as \emph{instruction tuning}, the LLM is finetuned on data of the type $I^\mathcal{T}, q^\mathcal{T}, a$ where $I^\mathcal{T}$ represents the instructions for a task $\mathcal{T}$ describing how the task should be performed, $q^\mathcal{T}$ represents a query from task $\mathcal{T}$ and $a$ represents the answer. The finetuning is performed on data from multiple tasks and multiple queries from each task. It is also possible to combine instructions with few-shot demonstrations in which case an example prompt may be as follows:
\[\mathrm{[Task~instructions]}\]
\[\mathrm{Demonstration~1:~} \mathrm{Query:}~q_1,~~\mathrm{Answer:}~a_1\]
\[...\]
\[\mathrm{Demonstration~k+1:~}\mathrm{Query:}~q_*,~~\mathrm{Answer:}\]

\paragraph{Benefits of ICL:}
Compared to the aforementioned approach of utilizing LLMs which involves pre-training followed by fine-tuning, ICL offers several key advantages. Firstly, fine-tuning may not always be feasible due to restricted access to the LLM, inadequate computational resources, or inadequately labeled data~\citep{brown2020language}, whereas ICL requires fewer resources, less data, and is easier to serve through API calls.
Additionally, ICL avoids the issues commonly associated with fine-tuning, such as overfitting~or~shocks~\citep{ying2019overview,kazemi2023understanding}, as it does not modify the model's parameters, allowing it to remain general.

\subsection{What Makes for Good Demonstrations?} \label{sec:good-demonstrations}
Several works try to provide theoretical justifications and insights into how LLMs learn from a few in-context demonstrations \citep{xie2021explanation,garg2022can,von2023transformers}. However, the exact reasons behind this capability are still largely unclear making it difficult to select optimal few-shot demonstrations. Fortunately, various empirical results show the effect of the few-shot demonstrations on the predictive accuracy of the LLMs and provide suggestions on the best practices for preparing them. They also show the brittleness of the LLMs in the choice, format, and order of the few-shot demonstrations. Here, we describe some of the more prominent ones.

\paragraph{Number of Demonstrations:} LLMs generally benefit from more demonstrations, but as the number of demonstrations increases the rate of improvement typically decreases \citep{brown2020language,ye2023context,min2022rethinking}. Generation tasks have been shown to benefit from an increased number of demonstrations more than classification tasks \citep{li2023unified}. Toward increasing the number of demonstrations, one barrier is the maximum context size of the LLM. While the size of the context has been increasing over time with newer LLMs, it may still be problematic for datasets with long input texts or classification datasets with many classes.

\paragraph{Demonstration Formatting:} Various works have shown that the formatting and wording of the prompts can play a crucial role in the performance of the LLM \citep{jiang2020can,shin2020autoprompt,kojima2205large,yang2023large}. For example, \cite{kojima2205large} show that simply adding \emph{Let's think step by step} to the prompt makes LLMs reason step by step and solve substantially more problems, and \cite{weller2023according} show that adding \emph{According to Wikipedia} to the prompt makes them more factual. Moreover, \cite{min2022rethinking} shows that besides the text formatting, the label space and the distribution of the input text in the demonstrations are also of immense importance.

\paragraph{Order of Demonstrations:} The order of demonstrations has been shown to substantially affect the model performance. For example, \cite{lu2022fantastically} show that on some tasks, the model performance can range from near-random to state-of-the-art depending on the order of the prompts, and \cite{zhao2021calibrate} show that answers appearing toward the end of the prompt are more likely to be predicted by the model. 

\paragraph{Diversity of Demonstrations:} Another important factor in the success of few-shot learning is the diversity of the demonstrations. \cite{naik2023diversity} propose \emph{DiversePrompting} where for the question of a demonstration, an LLM is used to generate different ways of solving the problem, and then those solutions are used in the prompt. \cite{zhang2022automatic} propose to select a diverse set of questions as few-shot examples. \cite{ma2023fairness} propose a fairness metric for selecting demonstrations which encourages selecting diverse few-shot demonstrations that produce a near uniform predictive distribution for a semantic-free input. 

\paragraph{Chain of Thought (CoT):} It has been shown that including a rationale for the answer significantly improves model performance, especially for models that are larger than a certain size \citep{suzgun2022challenging}. The rationale is commonly known as \emph{chain of thought (CoT)} \citep{wei2022chain}. In the case of CoT prompting, the demonstrations are typically formatted as:
\[\mathrm{Query:}~q_i,~~\mathrm{Rationale:}~r_i,~~\mathrm{Answer:}~a_i\]
with the rationale appearing before the final answer. Several works have investigated the reason behind the efficacy of CoT prompting and how to improve the prompts and rationales \citep{wang2022towards,lanham2023measuring}.

\section{In-context Learning with Demonstration Retrieval}
Traditionally, the same set of few-shot demonstrations is used on all queries, which can be suboptimal especially when there are high variations among the queries. An alternative is to \emph{retrieve} few-shot demonstrations that are tailored to the current query. Previous work has shown that demonstration retrieval leads to substantial improvements in the task metrics, compared to manually curated or randomly selected demonstrations \citep{luo2023dr,ye2023compositional}. Furthermore, LLMs have been shown to become less sensitive to the factors such as demonstration ordering (Section~\ref{sec:good-demonstrations}) when retrieved demonstrations are used \citep{li2023unified}.

This section gives an overview of the retrieval-based ICL (\reticl). We start by defining ICL with \emph{retrieved demonstrations}.
Formally, given a query $q_*$ and a \textbf{retrieval corpus $\mathcal{C}$}, a \textbf{demonstration retriever} $\mathcal{DR}$  selects a set of demonstrations $\{d_1, \dots, d_k\} \sim \mathcal{C}$, where each demonstration is $d_i = (q_i, a_i)$. The LLM input sequence becomes $(d_1, \dots, d_k, q_*)$.
The goal of the retriever is to select demonstrations that maximize the probability of the correct answer $a_*$.

The success of \reticl\ depends on several factors. This section explores design choices, including the retrieval objectives, retrieval inference strategy, and retrieval corpus. Then in Sections~\ref{sec:off-the-shelf}~and~\ref{sec:ft-retriever}, we explore the retriever models and how to train them to tailor to downstream tasks.

\subsection{Retrieval Objectives: Similarity and Diversity}
Various retrieval objectives for selecting and tailoring in-context examples for LLMs have been explored~\citep{luo2023dr, rubin2021learning, ye2023compositional, dalvi2022towards, cheng2023uprise, li2023unified}. 
There are two primary retrieval objectives for selecting demonstrations: similarity and diversity. 
Similarity involves selecting demonstrations most akin to the query and can be based on language similarity (term matching or semantic matching), structural aspects (sentence structure, reasoning structure, etc.), or other criteria. 
Most studies focus on language similarity, with fewer addressing structural similarity, often due to the challenges in extracting a query's structure in many tasks~\citep{levy2022diverse}.
Beyond similarity, some work has found that the diversity of demonstrations is important. The motivations for diversity include avoiding repetitive demonstrations~\citep{zhang2022automatic}, bringing different perspectives~\citep{yu2023generate}, and maximizing the demonstrations' coverage of the test query, in terms of covering either its words or syntactic structures~\citep{levy2022diverse}. Measuring the diversity of multiple demonstrations is a major technical challenge. 
~\citet{ye2023compositional} applied
determinantal point processes (DPP)  a probabilistic model to measure the negative interaction~\citep{kulesza2012determinantal}, to measure the diversity. 
~\cite{levy2022diverse} found that diversity and coverage are important when the model is unfamiliar with the output symbols space. 
It is noteworthy that researchers have found that ICL benefits more from demonstrations with higher complexity in some scenarios~\citep {fu2022complexity}, where they define the complexity in terms of the query length or reasoning steps. 
However, ~\citet{fu2022complexity} employed heuristic rules to define complexity and pre-selected demonstrations accordingly. Their research revealed that using a similarity-based retriever led to improved performance in a specific mathematical reasoning task. This might indicate that combining similarity and complexity considerations could be a promising strategy for enhancing the approach to reasoning tasks.  

\subsection{Inference Strategy to Retrieve Few-shots Demonstrations} \label{sec:retrieval-inference-strategy}
This section explores various strategies for employing a retriever to gather $k$ demonstrations. We divide these into three distinct methodologies.

\paragraph{One-hoc Retrieval}
This is the most basic retrieval strategy. 
To obtain $k$ demonstrations, given a query, the retriever ranks the demonstrations based on some scoring criteria and then selects the top-$k$ demonstrations. Thus, each demonstration is chosen independently of the others. This method is straightforward and fast, however, it might not yield the best combination of $k$ demonstrations as these demonstrations might be homogeneous.  
% \paragraph{One-hoc Retrieval}
% This is the most basic retrieval strategy to obtain $k$ demonstrations. Typically, this involves a retriever selecting $k$ demonstrations from a single corpus in one retrieval step. This is one of the most widely adopted retrieval pipelines used in the current Ret-ICL~\cite {}. While being simple, such a pipeline might retrieve multiple similar demonstrations which might not yield the best combination of $k$ demonstrations. 
% \xx{"Z-ICL: Zero-Shot In-Context Learning with
% Pseudo-Demonstrations" also talk about "diverse nearest" as a strategy besides KNN, which may be briefly mentioned in this section as well.}

\paragraph{Clustering Retrieval} 
To mitigate the issue of homogeneity in one-hot retrieval, clustering retrieval approaches ~\citep{li2022self,zhang2022automatic,li2023mot} categorize all demonstrations into $k$ sub-groups aiming to group similar demonstrations together. Then given a query, the retriever picks the most similar demonstration from each sub-group resulting in a final set of $k$ demonstrations. 
The core principle of clustering is to select a diverse range of demonstrations. Most of the work use SBERT~\cite{Reimers2019sbert} to encode the demonstrations (only the question or the entire demonstrations) and then apply $k$-means for clustering.

\paragraph{Iterative Retrieval}
The earlier retrieval strategies acquire each demonstration independently. However, in iterative retrieval, a retriever selects demonstrations based on both the query and previously retrieved demonstrations. This process starts with a single query, for which the retriever finds one best demonstration. The query is then augmented (e.g. combined with the demonstration)  to retrieve the next demonstration. This step is iteratively executed $k$ times to gather $k$ demonstrations. The general idea is to select the demonstrations that can complement each other. 
An an example of a work from this categorym, \citet{scarlatos2023reticl} train an LSTM retriever using a reinforcement learning framework. 
During the inference phase, the retriever processes the input query to select the best initial demonstration. It then generates a new query representation by integrating the query with prior demonstrations, specifically utilizing the hidden state representation from the LSTM model. This process of updating the query representation and obtaining subsequent demonstrations continues iteratively until $k$ demonstrations are retrieved.  

% \paragraph{Structure Search} For logical-form documents, \emph{encoding syntactic structures} of the documents and using their similarities for retrieval has successful applications in semantic parsing tasks \citep{levy2022diverse}. \ml{Do we have more in this category?}

\subsection{Retrieval Corpus} \label{sec:retrieval-source}

% \ml{the name should be consistent with the table 1}
The retrieval corpus forms a pool of demonstrations that the retriever can access.
Using annotated data is one of the most straightforward ways to construct the retrieval corpus. This setting assumes that training data related to a task is available, and thus can be used as the retrieval corpus. Under this setting, there are three main ways to construct the corpus that we will discuss individually below. 

\paragraph{In-Domain}
In this setting, an in-domain training set, independently and identically distribution (IID) with the test queries, is available and serves as the retrieval corpus. 
Most existing work take the full training set as the corpus.
However, to be more annotation efficient, \cite{hongjin2022selective} uses only a subset $M$ of the training set $N$ which includes the most representative and diverse ones, where $|M|<<|N|$. 
One question that remains unanswered from the work of \cite{hongjin2022selective} is how the predictive performance is affected as a function of retrieving from a subset $M$ instead of the entire training set $N$. While there is no follow-up work to answer this question, the closest comparison we find is the results in ~\cite{ye2023compositional} where a similar setup as \citet{hongjin2022selective} is used except that they use the entire training set as the retrieval corpus, and report lower performance on the SST-5 dataset (compare the Figure 3 in \cite{hongjin2022selective} and Table 3 in \citep{ye2023compositional}).  
While there might be other differences between the two setups that may affect the final performance, this comparison implies that retrieving from a carefully selected subset might have comparable results to retrieving from the entire training set.
% Both ~\citet{li2023mot,zhang2022automatic} assume access to unlabelled in-domain data and use LLMs to generate multiple CoT-answer pairs for such data. 
% \citet{zhang2022automatic} does not select verified if the generated CoT is correct but they found that selecting diverse demonstrations is important. 
% On the other hand, ~\citet{li2023mot} selects high-quality answers by using self-consistency (majority-generated answer). They repeat this process for each unlabelled data and combine them to form a retrieval corpus.   

\paragraph{Mix-Domain}
The previous scenario has one individual retrieval corpus for different tasks. Assuming that we want to test model performance on two tasks, then in the in-domain setting, there will be two retrieval corpora separately. Furthermore, the in-domain setting assumes that the model has knowledge about which task the test question belongs to such that when it comes to the retrieval phase, it knows which corpus to select the demonstrations from. However, this assumption does not hold in several real-world applications of LLMs. In the mix-domain setting~\citep{wang2023learning,li2023unified}, the retrieval corpus is constructed from the combination of all tasks. 
At the inference time, given a question, the retriever will retrieve demonstrations from this mixed corpus; the demonstrations can come from the same domain as the test question or from other tasks. 

\paragraph{Cross-Domain}
In this setting, IID human-annotated demonstrations are not available for the test queries, so one uses annotated demonstrations from other similar tasks~\citep{cheng2023uprise,shi2022xricl}. Note that this is different from the mix-domain setting where part of the corpus is IID and part of it is not. 
For instance, \citet{shi2022xricl} describes a scenario where the goal is to parse a Chinese query into SQL. However, the demonstrations are sourced from an English Text-to-SQL corpus, a domain with significantly more resources than the target domain. \citet{shi2022xricl} employs this high-resource data as the retrieval corpus. To adapt to the target domain during inference with a LLM, the target query is translated into the same language as the demonstrations. \citet{Nie2022CrossLingualRA} presents a similar approach, retrieving demonstrations from high-resource domains to address low-resource queries. However, their retrieval pool consists of multiple high-resource sources.

\paragraph{Unlabelled Queries with Automatically Generated Answers} The previous three corpora all presuppose the availability of human-annotated data. However, this assumption may not hold in real-life scenarios, particularly in streaming settings where users can pose questions without any pre-annotated answers. Several studies~\citep{zhang2022automatic,li2023mot} have suggested using LLMs to generate answers for unlabeled data. 
They apply filtering techniques to determine the quality of these generated answers, adding only those examples with high-quality answers to the retrieval corpus. The most widely used filtering technique is based on self-consistency~\citep{wang2022self}. This approach involves prompting the language model to generate multiple chains of thought and answers, then selecting the most common answer as the final response.

\paragraph{Free Form Corpus}  
Another approach to deal with the lack of human-annotated data for similar tasks is create pseudo-demonstrations from unstructured text. Toward this goal, \citet{lyu2022z} utilized the Demix dataset~\citep{gururangan2022demix}, which is not tailored for any specific task. To generate pusedo-demonstrations, a retriever selects the top-k most relevant sentences from the dataset. 
Subsequently, arbitrary labels are attached to each sentence to form the examples. 
~\citet{li2022self} propose a synthetic question answering generation method to create QA pairs using the synthetic generated passages by an LLM.
%To circumvent the issue of the model mimicking the answers from the most closely related examples provided in the context (also known as copy issue of ICL), sentences that are close to but not exactly the retrieved ones are used, along with synonyms for the task labels.

% % The foundation of these methods is the demonstration retriever. A typical demonstration retriever encodes examples from a pre-cached demonstration pool and the query into some vector representations, and then a similarity measure (e.g. cosine similarity) is calculated between candidate demonstration embeddings and the query embedding to locate the most related demonstrations. An effective retriever empowers researchers to design targeted retrieval strategies for different downstream tasks, contexts, and requirements.

% Given the limited understanding of the underlying mechanism through which retrieved demonstrations enhance the performance of large language models, initial research efforts focused on a heuristic evaluation of readily available retrievers for this task. Subsequent research endeavors explored the design and development of learning-based retrievers specifically customized for the purpose of retrieving demonstrations. This section reviews representative models from both approaches.

% \subsection{Off-the-Shelf Retriever}
% Recent research has demonstrated success of using off-the-shelf retrieval methods to help improve the performance of LLMs.  \ml{add some citations here}

\section{Off-the-shelf Demonstration Retrievers}\label{sec:off-the-shelf}

To achieve the retrieval objectives outlined above, researchers have explored various types of demonstration retrievers. A typical demonstration retriever encodes examples from the retrieval corpus and the query into some vector representations, and then a similarity measure (e.g. cosine similarity) is calculated between candidate demonstration embeddings and the query embedding to locate the most related demonstrations. Given the limited understanding of the underlying mechanism through which retrieved demonstrations enhance the performance of LLMs, initial research efforts focused on a heuristic evaluation of readily available retrievers for this task. Subsequent research endeavors explored the design and development of learning-based retrievers specifically customized for retrieving demonstrations. This section reviews representative off-the-shelf models and we will discuss the learning-based models in Section \ref{sec:ft-retriever}.
 
\paragraph{Term-based Similarity} BM25~\citep{robertson2009probabilistic}
% , a widely-used SR method, was introduced by Robertson and Walker in 1994 as an improvement over the previous Okapi BM11 algorithm. 
is one of the most popular term-based scoring methods due to its simplicity and effectiveness in producing relevant results. It takes into account both term frequencies and document lengths.
% It is a family of scoring functions that takes into account both term frequency and document lengths normalization to determine the relevance of the documents to a given query. 
% In the literature, the most widely-used heuristic fewshots selection strategy is to use BM25 to select most similar examples in terms of corpus overlap for input queries, and it has been empirically shown to improve prediction performance on many different tasks
It has been empirically demonstrated in various works \citep{luo2023dr, rubin2021learning, agrawal2022context, ye2023compositional, dalvi2022towards} that using BM25 to select similar examples as few-shots in ICL can help improve the performance of many LLM inference tasks. While BM25 has become a standard baseline model in the field, it is not without its limitations. Due to its sole reliance on term frequency and document length, this approach may overlook crucial aspects such as semantic meaning and sentence structure, potentially leading to inaccuracies in certain instances. 
Another drawback is that BM25 lacks the capability for fine-tuning in downstream tasks, making it less competitive compared to neural models which can be fine-tuned and customized for specific downstream tasks.

% \xx{Move this paper to strategy section: \cite{levy2022diverse} claimed that selecting top-k most similar demonstrations using BM25 is not always the optimal method under some circumstances: they justified it in the compositional generalization setup (where train and test programs do not overlap) for semantic parsing. In this case, combining diverse demonstrations retrieved from BM25 retrievers that cover different local structures outperforms top k retrieval in both w/o finetuning setups.}

\paragraph{Sentence Embedding Similarity}
% One promising DR solution is to utilize sentence embedding similarity technique. 
In this approach, queries and documents are encoded to the same dense embedding space using an off-the-shelf sentence embedding model, and then similarity scores (e.g. cosine similarity) are calculated to rank the most relevant documents for each query. A rich collection of sentence embedding methodologies exists in the literature.
Sentence-BERT (SBERT)~\citep{Reimers2019sbert} is a modification of the pretrained BERT network that uses siamese and triplet network structures to derive semantically meaningful sentence embeddings. The effectiveness of SBERT embeddings for demonstration retrieval has been investigated in several works \citep{rubin2021learning, li2023mot, wang2023learning}, and the results show that retrieving demonstrations based on SBERT embeddings often provides a boost in performance compared to zero-shot or random few-shot selection. In the KATE method \citep{liu2022makes}, the authors studied using vanilla RoBERTa~\citep{liu2019roberta} and finetuned RoBERTa on NLI~\citep{Bowman2015snli} and STS-B~\citep{cer2017stsb} datasets for selecting good demonstrations, and found that the finetuned version on task-related datasets offered further empirical
gains. Note that here, the demonstration retriever is not trained for ICL demonstration retrieval based on task-specific data (a topic which we will discuss in Section \ref{sec:ft-retriever}); instead, the retriever is finetuned related tasks to provide a better notion of similarity for the task at hand. So we still categorize it as an off-the-shelf retriever. \cite{shi2022xricl} extends the use case to cross-lingual few-shot retrieval in the Text to-SQL semantic parsing task, and they use mSBERT~\citep{reimers2019sentence}, mUSE~\citep{yang2019muse} and mT5~\citep{xue2020mt5} as the baseline models for comparison. Other widely used baseline models for demonstration retrieval include $E5_{\text{base}}$~\citep{wang2022text}, SimCSE~\citep{gao2021simcse}. Instead of relying on “word matches” as in BM25, these sentence embedding similarity approaches can better capture semantic similarity (for example, synonyms, and related topics), however computationally they might be more expensive.
% ~\citet{li2023mot} firstly uses SBERT to find the top-k demonstration candidates from each cluster, and then prompts the LLM to select which demonstration is most helpful for answering the target question. \xx{This last paper is already mentioned in the SBERT part, to retrieve from different clusters I think should be the inference strategy, rather than mentioned here.}

\paragraph{Pretrained Dual Encoder}
% Compared with previously discussed retrievers like SBERT, LLM retrieval architecture like Dual Encoder can better capture the deep connection of complicated logic and reasoning than applying same semantic embeddings between query and candidates~\citep{li2023mot}. 
In the context of demonstration retrieval where the goal is to identify relevant examples for a given query, the query is typically a question, while the examples may contain additional information such as answers, chains of thoughts, supporting knowledge, or even follow different patterns. Therefore, transforming them into a uniform embedding space to calculate relevance might not be the most effective approach.
% Unlike previously discussed retrievers like SBERT, 
In this case, LLM retrieval architectures such as Dual Encoder that are pretrained on retrieval or question-answering tasks can better grasp the intricate relationships between complex logical concepts and reasoning processes by employing different semantic embeddings for queries and candidates~\citep{li2023mot}. 
% Particularly in the demonstration retrieval scenario, we would like to find out good examples for a particular query, where the query is typically a question, while the examplers might come with additional information such as answer and chain-of-thought, or even in different pattern, so transforming them into the same embedding space to calcuate the relavance might not be the best solution.
% It works by producing close dense vectors for queries and documents in the embedding space if they are considered relevant. 
In practice, training a dual-encoder can be highly expensive as it typically requires a large training corpus. Fortunately, there are publicly available pretrained retrievers, although not specifically optimized for few-shot retrieval tasks, already demonstrating success in helping LLMs to learn from the selected examples.
\cite{luo2023dr} studied applying GTR~\citep{ni2021large} to select semantically similar examples as demonstrations, and empirically proved that this approach brought in better performance gain than random fewshots for both PaLM~\citep{chowdhery2023palm} and FLAN~\citep{chung2022scaling} models. 
%Here, GTR(Generalizable T5-based dense Retrievers) was introduced by \cite{ni2021large}. 
GTR is a T5-based dual encoder model that is pretrained on the CommunityQA~\citep{abujabal2019comqa} and finetuned on the MS Marco dataset~\citep{nguyen2016ms}. Moreover, \cite{khattab2022demonstrate} reported results for employing ColBERTv2~\citep{santhanam2021colbertv2} as the retrieval module in their DEMONSTRATE–SEARCH–PREDICT (DSP) framework for ICL. ColBERTv2 is a state-of-art retrieval model that adopts the late interaction architecture~\citep{khattab2020colbert} and is trained on the MS Marco dataset. In the proposed framework, it is used to retrieve both (i) related knowledge during the search stage and (2) top k similar examples as demonstrations.

\section{Fine-tuned Demonstrations Retrievers}\label{sec:ft-retriever}

% \yl{My two cents: it seems that whether using feedback signals from LLM is an important difference among the different training dataset/method.}
% \ml{Architecture-wise, we do not observe the difference of demonstrations retriever v.s. general information retrievers such as those applied for the open-domain question answering. There are two types of fine-tuned retrievers: bi-encoder retriever and cross-encoder reranker. 
% It is better to have a figure to show the difference of two. 
% On the other hand, the general retrievers are trained on the annotated question-answering pairs, for the demonstrations retrievers, there are no available annotated data, thus, a challenge is how to get the training data.  There are following approaches in general.}

Although off-the-shelf retrievers have shown some promise in retrieving demonstrations for LLMs, the retrieved demonstrations given by the off-the-shelf retrievers might not represent the nature of the task and how the task should be solved in general. Therefore, it might lead to sub-optimal performance. 
Researchers thus have started to explore learning-based methods to further push the boundaries. A typical objective when designing a good demonstration retriever is: if an LLM finds a demonstration useful when being used as an illustrative example, the retriever should be encouraged to rank the demonstration higher. This allows us to train models directly relying on signals from query and output pairs in the task of interest, without human annotations. 
% On the other hand, demonstration retrievers share a lot of common ground with traditional information retrievers and therefore can exploit the existing architectural advances with some tweaks to training data or objective functions. Based on these observations, researchers designed various finetuned demonstration retrievers that are guided by task supervision.
% and improve the demonstration selection capabilities for LLMs.
To develop a demonstration retriever, the majority of approaches utilize current dual encoder models~\citep{karpukhin2020dense,ni2021large}. The key variations lie in the methods of gathering training data and formulating training objectives. We will explore these aspects in more detail in the subsequent sections.

\subsection{Collecting Training Data for Demonstration Retriever}
% \paragraph{Target Similarity} 

\paragraph{Based on LLMs Signals}
% \paragraph{Rank Candidates Based on Conditional Probability}
A popular approach to collecting training examples is to use the supervisory signals from LLMs. In this case, a typical paradigm is to first employ some filtering mechanisms~\citep{cheng2023uprise} or unsupervised retrievers (e.g. BM25 and SBERT)~\citep{luo2023dr} as the initial retriever, this step can help limit the pool size for mining the right training data. Then a scoring LLM, which serves as a proxy for the inference LLM, is used to score each candidate demonstration $d$.
% ~\citep{rubin2021learning, luo2023dr, wang2023learning, li2023unified}. 
Here the score is defined as $s(e) = p(a|d, q)$ which is the conditional probability of output answer $a$ given the input query $q$ and demonstration $d$. Another approach is to train a smaller reward model that can provide more fine-grained supervision for dense retrievers. For example, \cite{wang2023learning} proposed to finetune a cross-encoder model serving as a teacher model for training the retriever. 

Once a score is obtained, a retriever can be trained that predicts these scores directly~\citep{ye2023compositional}. Alternatively, the candidate demonstrations can be ranked for each query based on their scores, considering the top-ranked demonstrations as \emph{positive} examples that help the LLM get to the right answer and the bottom-ranked ones as \emph{negative} examples that mislead the LLM towards the wrong answers; then a retriever can be trained which separates positive examples from negative examples~\citep{rubin2021learning,cheng2023uprise,luo2023dr}. 

There are different strategies for choosing the scoring LLM. Ideally, one uses the inference LLM itself as the scorer in order to perfectly reflect its preferences~\citep{li2023unified, shi2022xricl}. However, training retrievers requires large amounts of labeled data, and it may be expensive use very large models for labeling. Consequently, for scoring one may gravitate towards utilizing smaller models, especially those within the same model family as the inference LLM~\citep{luo2023dr, cheng2023uprise, rubin2021learning}. 
% For example, \cite{luo2023dr} employs a 62b PALM model as scoring model while a 540b PALM model is used for inference. 
% \cite{rubin2021learning} use GPT-NEO, a 2.7B-parameter LM, serving as a scoring model when applying GPT-J(6B parameters), GPT-3(175B parameters) and CODEX(175B parameters) as inference model. Similarly, \cite{cheng2023uprise} scores the examples using GPT-Neo for inference tasks on BLOOM(7.1B parameters), OPT(66B parameters) and GPT3(175B parameters). 

% \paragraph{Rank Candidates using Trained Reward Model}

% \xx{Integrate Yue's writing: Detailed methods of this strategy vary. \cite{li2023finding} designed an \textit{InfoScore} for each demonstration $e$, which essentially aggregates $Prob_{G}(\bar{y_i}|e,\bar{x_i})$ across a representative subset $\{\bar{x_i},\bar{y_i}\}\subset D$, and use that score for retrieval. The most common method is to use the scoring model to create a retriever training set $\{\bar{x_i}, e_j, Prob_{G}(\bar{y_i}|e_j,\bar{x_i})\}$, where both $(\bar{x_i},\bar{y_i})$ and $e_j$ are drawn from the training set $D$, and fine-tune a demonstration retriever that learns to measure the similarity between each test query $x$ and demonstration $e$, roughly in accordance to $Prob_{G}(y|e,x)$, so that best performing demonstrations can be retrieved~\citep{rubin2021learning,li2023unified,cheng2023uprise,shi2022xricl,luo2023dr}. We will discuss this family of methods extensively in section \ref{ft-retriever}, introducing different training objectives and methods. 
% }
\paragraph{Model-Free} 
% \ml{\citet{hu2022context} measures the similarity between the target of the query with each example to select the positive examples.}
One approach to collecting training data for demonstration retriever is to directly measure the similarity between the labels of the candidate demonstrations and the label of the query, and use this similarity as a proxy of the importance of a demonstration~\citep{hu2022context,poesia2021synchromesh}. 
For instance, \citet{hu2022context} explored a dialogue context where labels are structured as a sequence of stages. The similarity between a query's label and a demonstration's label is determined by calculating the average F1 scores of these two labels.
This method adopts a heuristic approach (i.e. stage changes), presuming that the similarity metric can closely resemble the preference for good demonstrations from an LLM, and it often necessitates domain-specific expertise for design.

\subsection{Training Objectives}
Thus far, we have explored the creation of training data for demonstration retrievers in the context of ICL. We now proceed to examine the commonly used loss functions for training retrievers.
% \paragraph{Binary Classification} used to train re-ranker. 

% \xx{\paragraph{Pair-wise ranking}? \cite{hu2022context} }
%\cite{hu2022context} select topk positive and negative examples, and use contrastive loss to optimize the retriever such that the score of the positive is higher than the negative. 
% In pair-wise ranking, the training data is composed of positive and negative pairs of data points. A positive pair consists of a query and a positive example that can help the LLM predict the correct answer, while a negative pair consists of the query and a negative example that might mislead the LLM. The list-wise ranking loss function is designed so that the similarity between a positive pair is high, while the similarity between a negative pair is low. More formally, we can write
% \[\mathcal{L}_{\text{listwise}}=
% \begin{cases}
% \text{sim}(x_i, e_i) & \text{if positive pair}\\
% max(m - \text{sim}(x_i, e_i)) & \text{if negative pair}
% \end{cases}
% \]

\paragraph{List-wise Ranking Loss}

The list-wise ranking approach looks at a list of candidate documents for a given query and tries to capture the correct ordering for it. 
% Inspired by LambdaRank~\citep{burges2010lambdamart}, \cite{li2023unified} proposed to inject the ranking signals into the retriever using a list-wise ranking loss. 
\cite{li2023unified} proposed to inject the ranking signals into the retriever using an approach inspired by LambdaRank~\citep{burges2010lambdamart}.
More formally, given each query $q$, they first rank all $l$ candidate documents according to their relevant scores $S=\{s(d_i)\}_{i=1}^l$, according to which the associated ranking $\mathcal{R}=\{r(d_i)\}_{i=1}^l$ is computed. 
% they first use the inference LLM to score each candidate example $e_i$, which results in a set of scores $\mathcal{S}=\{s(e_i)\}_{i=1}^l$. According to $\mathcal{S}$, they rank all the candidates and get their ranking $\mathcal{R}=\{r(e_i)\}_{i=1}^l$. 
Then the loss function is defined as follows:
\[\mathcal{L}_{\text{listwise}}=\sum_{d_i, d_j} \text{max}\left(0, \frac{1}{r(d_i)}-\frac{1}{r(d_j)}\right) * \log(1 + e^{\text{sim}(q_i, d_j)-\text{sim}(q_i, d_i)})\]
where $\text{sim}(q, d)$ is the relavance between a candidate demonstration $d$ and the input $q$. 
% One thing worth mentioning is that in order to fully leverage the computation of the same batch, they also combine the in-batch negative loss 
% \[\mathcal{L}_{\text{in-batch}}=-\log{\frac{e^{\text{sim}(x_i, e^*)}}{\sum_{e\in E}e^{\text{sim}(x_i, e)}}}\]
% with $e^*$ as rank-1 candidate of $x_i$, and $E$ is all candidates in the current batch.
In the list-wise ranking objective, retriever can benefit from the full ranking of the candidate set to make accurate predictions for the most relevant demonstrations. However, obtaining the full ranking list and calculating the loss function on top of it might be very expensive and time-consuming. Additionally, the model is trained to discern the relative preferences between examples without explicitly determining whether an example can serve as an absolute good demonstration.

\paragraph{InfoNCE Loss}
Another widely adopted training procedure is contrastive learning using the InfoNCE loss~\citep{rubin2021learning,cheng2023uprise,luo2023dr}. When positive and negative examples can be correctly identified, InfoNCE loss is an effective loss function because it can take advantage of the supervisory labels to produce a representation that sets apart the useful examples for demonstration retrieval. In this approach, each training instance is given in the form of $<q_i, d_i^+, d_{i,1}^-,...d_{i,k}^->$. Here $d_i^+$ is a selected positive example concerning the input $q_i$, and the negative examples consist of one hard negative example $d_{i,1}^-$ and $k$ random examples from the other instances in the same mini-batch. Then the typical contrastive loss can be defined as 
\[\mathcal{L}_{\text{cont}}=\mathcal{L}(q_i, d_i^+, d_{i,1}^-,...d_{i,k}^-)=-\log\frac{e^{\text{sim}(q_i, d_i^+)}}{e^{\text{sim}(q_i, d_i^+)}+\sum_{j=1}^ke^{\text{sim}(q_i, d_{i,j}^-)}}\]
The random negative examples from the same mini-batch are called \emph{in-batch negatives}. They are typically selected from both the positive examples and hard negative examples of other instances.

% \textbf{In-batch Negative}
% In this process, negative examples are selected from within the same mini-batch of data that are using for training, instead of a set of pre-selected negative examples.
% This means that for each anchor example in the batch, other examples from the same batch are picked as the negative examples. 
% These examples are part of the current training batch and are readily available for computing gradients during training.

\paragraph{Distillation by KL Divergence} 
\cite{ye2023compositional} claims that although the InfoNCE loss has been found effective in training demonstration retrievers and can learn which examples might be superior to others, it has the same treatment for all negative examples and the predicted scores from LLM are not fully utilized.
As an alternative to train a demonstration retriever using positive and negative examples, \cite{shi2022xricl} proposed to train the retriever by directly distilling the LLM’s scoring function. More specifically, the retriever model is designed to produce ranking scores that match the usefulness of a demonstration to help with the LLM inference; this is done by minimizing the KL-divergence between the top $K$ examples score distribution from scoring LLM and the ranking score distribution produced by the retriever
\[\mathcal{L}_{\text{distill}} = \text{KL}(p_{\text{LLM}}||p_{\text{retriever}}) = \sum_{k=1}^Kp_{\text{LLM}}(d_k)log\left(\frac{p_{\text{LLM}}(d_k)}{p_{\text{retriever}}(d_k)}\right)\]
% Note that this loss is only used to optimize the parameters of the retriever, and not the language model. When using recent deep learning frameworks, this is achieved by applying a \emph{stop gradient} operator[] on $p_{\text{LLM}}$, which is defined as identity at forward computation time and has zero partial derivatives.

\paragraph{Multiple Objectives} In \cite{wang2023learning}, the authors proposed to train the demonstration retriever model with combined objectives: (1) knowledge distillation from the trained reward model which can capture the preferences of LLMs over the retrieved candidates (2) InfoNCE-based contrastive loss to incorporate the in-batch negatives. More specifically, the resulting loss function is as follows:
\[\mathcal{L}_{\text{combined}}=\alpha\mathcal{L}_{\text{cont}}+\mathcal{L}_{\text{distill}}\]
Here
%$\mathcal{L}_{\text{distill}}$ is the KL-divergence between the retriever and a finetuned reward model that can mimic the preference of the inference LLM, and
$\alpha$ is a constant that controls the relative importance of the two losses.
% In their model, the retriever is initialized with $\text{E5}_{base}$~\citep{wang2022text}.
They claimed that with the multi-objective function, both the absolute scores and supervised signals are taken into consideration.
\citet{li2023unified} trains a universal retriever with both list-wise ranking loss and the InfoCNE loss.

% \xx{Compare these objective functions
% Robustness, advantage in demonstration retrieval, computational cost, performance on different dataset, a table?
% }

% \paragraph{some other method}
% \cite{ye2023compositional} claims that although the InfoNCE loss has been found effective to train demonstration retrievers that can learn which exampler might be superior to others, it has the same treatment for all negative examples and the predicted scores from LLM are not fully utilized. Instead, they proposed to employ a fine-grained pair-wise margin loss to train retrievers. Here the loss function for each training instance $x_i$ is defined as
% \[\mathcal{L} = \]
 
% In other words, the first stage retriever performs top-k document retrieval, i.e. the potential set of documents relevant to the query; the second (and, in case, its successors) stage reranker aims at reordering that set of candidates with more powerful and computationally expensive models.

% \begin{figure} [H]
%     \centering
%     \includegraphics[width=\linewidth,trim={0 60 0 100},clip]{retriever-reranker.pdf}
%     \caption{Retriever-reranker architecture}
%     \label{fig:retriever-reranker}
% \end{figure}

\paragraph{Iterative Training} 
% \ml{It is better to move this paragraph before the retriever-reranker.}
Regarding training strategies, most research efforts have centered on fine-tuning a single retriever. \citet{wang2023learning} and \cite{li2023unified} instead proposed to iterate the retriever model multiple times. More specifically, the retriever trained in iteration $i$ will be employed to retrieve a new set of candidates for the subsequent iteration $i + 1$. Such an iterative training approach allows progressively improving retriever quality by mining better positive and hard negative examples at each iteration.

\paragraph{Diversity Training}
The Determinantal Point Process model~\citep{kulesza2012dpp} defines a probability distribution over all the combinations of candidate demonstrations, giving high probability to subsets that contain relevant and diverse items~\citep{levy2022diverse}. It models diversity by incorporating cross-candidate similarity scores, and models similarity via a per-candidate relevance score, i.e., a similarity score between a candidate and the test query. In addition to using DPP directly~\citep{levy2022diverse}, ~\cite{ye2023compositional} also fine-tuned a DPP model and demonstrated meaningful improvements over pure similarity-based methods.

\paragraph{Re-ranker Training}
It is not uncommon that people adopt a two-stage retriever-reranker architecture for ICL retrieval in the literature to further improve the exemplar selection process~\citep{shi2022xricl}. Generally, a dual-encoder-based retriever can encode query and candidate documents for fast indexing and searching, but neglect the finer-grained token-level interactions. Cross-encoder-based reranker, on the other hand, can capture the subtle relationship but is time-consuming. We can benefit from both of these methods by chaining two methods together. In the first stage, a retriever model is used to quickly select the top $N$ examplers to limit the candidate pool of interest, then a  reranker reranks the retrieved $N$ examplars and uses the top $K$ exemplars to construct a prompt. Sigmoid cross-entropy loss is typically used for training the reranker. 
~\citet{lu2022dynamic} also utilizes a similar structure as the reranker to select the demonstrations from random candidates. The reranker is trained using reinforcement learning. 

% \ml{Add equations of DPP}

\subsection{Summary}
% To summarize, we outline the advantages and disadvantages of various retriever models in Tabel~\ref{tab:pros_cons}.
Here, we summarize the advantages and disadvantages of various retriever models. 
The off-the-shelf retrievers are easy to use without any downstream task finetuning and typically demonstrate stronger performance than random demonstrations. One exception is in commonsense reasoning tasks where~\cite{zhang2022automatic} and \cite{ye2023compositional} found that for these tasks, random demonstrations are consistently better than retrieval-based method. \citet{cheng2023uprise} also show that retrieved demonstrations harm commonsense reasoning and coreference resolution tasks. Among the three categories of off-the-shelf retrievers, sparse retrievers such as BM25 are more index-efficient. This feature becomes particularly valuable when dealing with large volumes of demonstrations and limited hardware memory, making BM25 a preferable choice under such circumstances. In contrast, sentence-embedding similarity-based methods and dual-encoder-based retrieval systems, which are trained on language tasks, excel in capturing more semantically focused retrieval. 
Regarding performance, ~\citet{luo2023dr} compared BM25 with dual encoder (GTR) across 5 tasks, and they found that the average performance of these two is very similar (within 0.5\% difference), and BM25 outperformed the dual encoder in some tasks and vice versa. In another study, \cite{ye2023compositional} observed a similar trend highlighting that no single retriever consistently outperforms others across different tasks. Both \cite{rubin2021learning} and \cite{li2023unified} found that BM25 is better than SBERT on semantic parsing tasks, while ~\cite{li2023unified} found that SBERT is better than BM25 on sentiment analysis tasks. 
Nevertheless, retrievers that are fine-tuned demonstrate superior performance compared to their off-the-shelf counterparts. The main drawback of fine-tuned retrievers lies in the high cost of obtaining training data. Additionally, the common practice of employing task-specific retrievers complicates the system and limits its generalizability. 
~\citet{li2023unified} proposed to train a universal retriever that shows stronger performance than task-specific demonstration retriever (e.g. EPR~\citep{rubin2021learning}) on most of the tasks. 
% \input{pros_cons}

% \paragraph{Compare performance}
% ~\citet{rubin2021learning} shows that with GPT3, BM25 is better than SBERT on three semantic parsing tasks. 

% ~\citet{ye2023compositional} shows that Random is better than any other methods even the fine-tuned retriever on the CMSQA dataset (commonsense reasoning).

% ~\citet{cheng2023uprise} shows that retrieved demonstrations harm commonsense reasoning and coreference resolution tasks. 

% ~\citet{li2023unified} have study multiple classification and generation tasks. For some tasks, the performance of BM25 and SBERT is close and not a single retriever is better than the other one, however, on semantic parsing tasks, BM25 is consistently better than SBERT, and SBERT is better than BM25 on the sentiment analysis task. Furthermore, the universal retrieval is better than task-specific demonstration retriever (e.g. EPR~\citep{}) on most of the tasks. 

% RetICL~\cite{}
% PROMPTPG~\cite{lu2022dynamic}

\section{Applications} \label{sec:application}

The effectiveness of retrieval-based ICL has been showed in four categories of tasks. 1). natural language understanding, 2- reasoning, 3- knowledge-based QA, and 4- Text generation. We discuss each category below.

Natural language understanding tasks that benefit from \reticl\ include sentiment analysis (SA)~\citep{socher2013recursive,zhang2015character,go2009twitter},
paraphrase detection (PD)~\citep{dolan2004unsupervised,paws2019naacl},
reading comprehension (RC)~\citep{rajpurkar2016squad,MultiRC2018,clark2019boolq,MultiRC2018,clark2019boolq,OpenBookQA2018}, and natural language inference (NLI)~\citep{williams2018broad,wang2018glue,bowman2015large,de2008finding}. 
Specially, \reticl\ shows noticeable improvements on SA and NLI tasks~\citep{liu2022makes,ye2023compositional}. 

Reasoning tasks tasks that benefit from \reticl\ include mathematical reasoning~\citep{cobbe2021gsm8k,lu2022dynamic,ling2017program}, commonsense reasoning (CSR)~\citep{talmor2019commonsenseqa,zellers2019hellaswag,Bisk2020,roemmele2011choice},and ethical Reasoning~\citep{jiang2021can}.  
%Coreference Resolution (CR)~\citep{levesque2012winograd}. 
Such tasks are usually accompanied by CoT, but ~\citet{zhang2022automatic} found that CoT does not help that much for the commonsense reasoning task. Many works have shown that retrieval-based methods are worse than random demonstrations on commonsense reasoning tasks (e.g. CMSQA). A simple similarity-based retrieval method does not show significant improvement in mathematical reasoning tasks, and ~\citet{zhang2022automatic} shows that diversity is important for mathematical reasoning tasks. The iterative retrieval strategy shows the most significant improvement on mathematical reasoning tasks~\cite{scarlatos2023reticl}.

In Knowledge-based QA, external knowledge is required to answer the question~\citep{berant2013semantic,kwiatkowski2019natural,joshi2017triviaqa,Clark2018ThinkYH}.
To tackle such tasks, the state-of-the-art systems usually retrieve relevant passages that might contain the answer to the question, and then feed such passages and questions together to a language model to generate the answer. 
\citet{liu2022makes} shows that using retrieval-based ICL (sentence semantic similarity-based retriever with GPT-3) is almost comparable to a fine-tuned method. 
\citet{ye2023compositional} shows that BM25 achieve 10+\% improvement on open-domain QA. 

Text generation tasks that benefit from \reticl\ includes code generation (CodeGen)~\citep{zelle1996learning,lin2018nl2bash}, semantic parsing (SP)~\citep{wolfson2020break,li2021mtop,andreas2020task}, 
text-to-SQL~\citep{shi2022xricl}, Table-to-text (Table2Text) generation~\citep{parikh2020totto}; 
Data-to-Text (D2T)~\citep{nan2021dart,duvsek2019semantic}.
~\citet{rubin2021learning} shows that the retrieved demonstrations significantly outperform random demonstrations (e.g. BM25 is 25+\% better than random, and EPR is 30\% better than random).  

Apart from different types of tasks, \citet{hongjin2022selective} shows that in scenarios with limited training data, RetICL outperforms fine-tuning a model on such sparse data. Furthermore, leveraging data from a high-resource domain can enhance performance in a low-resource domain, as seen in cross-lingual contexts~\citep{shi2022xricl,Nie2022CrossLingualRA,cheng2023uprise}. 

% \mk{I suggested that we look for papers that are \textbf{not} specifically focused on demonstration retrieval, but they try to solve some task X and they find that adding demonstration retrieval helps improve their task. For example, we may find a paper that tries to solve molecule generation with LLMs, and they mention that demonstration retrieval helped them.}

\section{Discussion of Future Direction}

\paragraph{Retrieve Demonstrations From Raw Text}
Much research assumes the availability of annotated samples that can be utilized as a retrieval corpus. Yet, when faced with a novel task, it is often the case that no such training dataset exists. While there are preliminary efforts to create pseudo demonstrations from open-ended corpora like Wikipedia~\citep{lyu2022z}, the proposed method is restricted to classification tasks and the label to the demonstrations are randomly assigned. 
A potential approach to obtain pseudo demonstrations for generation tasks is ~\citet{wan2023better}, where they assume a set of unlabelled queries available (without ground truth labels), and use LLMs to generate chain-of-thoughts and answers and then apply self-consistency~\citep{wang2022self} to select high-quality demonstrations to form pseudo demonstrations pool. 
Employing this method of generating answers with sentences retrieved from a free-form corpus could potentially create high-quality pseudo demonstrations.

% The generation task is a more flexible format and thus has more applications in the real world, but constructing the demonstrations of the generation task is still an open question.

% \paragraph{Demonstration Retriever Training and Selection} 
% In Section~\ref{sec:ft-retriever}, we have summarized different ways of training a retriever to search the demonstrations. 
% Most of the methods depend on using an LLM to select the positive and negative demonstrations of a question to train a retriever.
% Such a paradigm not only requires lots of inference computation, but also it is unclear which signal should be used to decide the positive and negative demonstrations which means that the training data of the retriever might not be of high quality~\citep{hashimoto2023take}.
% What type of retriever to use is another question, for example, choosing a neural model retriever, sparse retriever, or template-based retriever? 
% Right now, what we have seen is that for different tasks, sparse and dense retrievers perform differently. It is unknown whether there will be a type of retriever that can consistently yield better performance than others.  

\paragraph{Choosing the Type of Retriever}
Another critical consideration in this domain is the selection of the type of retriever. The options range from neural model retrievers and sparse retrievers to template-based retrievers. The objectives range from similarity and diversity to complexity. 
Current research implies that no single type has emerged as universally superior. This leads to an important open question: Is there a potential for a specific type of retriever to consistently yield superior performance across a variety of tasks? Investigating this will be a key direction for future research.

\paragraph{Retriever Training Methods}
In Section~\ref{sec:ft-retriever}, we explore various methods for training a retriever to search demonstrations. These methods largely depend on using an LLM to identify positive and negative demonstrations for a given question. This approach, while innovative, comes with significant computational demands. Moreover, the ambiguity in choosing what constitutes a positive or negative demonstration raises concerns about the quality of the training data for the retriever~\citep{hashimoto2023take}. Addressing these challenges is crucial for the development of more efficient and reliable retriever training methods.

\paragraph{Active Demonstration Retrieval}
% \mk{Maybe relevant: https://arxiv.org/pdf/2302.12246.pdf}
Much of the current research is based on a static framework where the retrieval corpus remains constant. In practical situations, input distributions may change over time and models may come across new instances. In these cases, one may like to keep updating the retrieval corpus based on the new incoming queries, but since the labels for the new samples may not be available, a selection strategy is needed to select a representative subset of the incoming examples for annotation so they can be added to the retrieval corpus. This problem is reminiscent of the well-studied active learning problem. 
~\citet{zhang2022active} have studied how to actively select examples with unlabelled data, however, this study is not based on the retrieval setting. The combination of active learning and \reticl\ can be an interesting future direction. 

\paragraph{Retrieved Demonstrations for Small LM} Most of the existing work focus on LLMs (more than 100 B parameters). Research into small LMs has recently received more attention due to its inference efficiency.
Examining and improving the ICL capabilities of small LMs by better (ideally optimal) demonstration selection is an interesting direction to explore.

\paragraph{Theoretical Understanding of Why Are Similar Demonstrations Better Demonstrations?}
While lots of research has shown that similar demonstrations are better than random, it is still unknown why similar demonstrations are helpful. 
~\cite{ye2023compositional} found that the generation task is more beneficial compared to the classification task, and one possible reason is that the retrieved demonstrations could have similar answers to the input question and the model might just copy the answers. However, such an explanation does not illustrate why retrieved demonstrations are better than random ones on classification tasks. 
There are some hypotheses that similar demonstrations help locate the knowledge in the LLMs or based on the hypothesis of LLM conduct implicit gradient descent, then the similar demonstrations provide a more useful training signal to the input query. The research on why ICL works can help understand why similar demonstrations are better than random ones. 

\paragraph{Fine-tuning For RetICL Few-shot Learning}
Most of the existing research utilizes a frozen LLM as the few-shot learner in the RetICL framework. As demonstrated by \cite{gao2021making,izacard2023atlas}, fine-tuning LLMs on a few-shot learning task can be a promising approach to enhancing LLM performance during inference. An intriguing avenue could involve adapting the fine-tuning strategy from open-domain question answering models to RetICL~\citep{lewis2020retrieval,guu2020retrieval}.

\paragraph{RetICL for Vision and Language Models} 
Vision and language models (VL) have demonstrated proficiency as few-shot learners \citep{alayrac2022flamingo,awadalla2023openflamingo,li2023mimic}, and researchers have integrated retrieval augmented generation methods with VL models \citep{luo2021weakly,Gao2022trig,yasunaga2023retrieval}. In lieu of employing random demonstrations, \cite{yang2024exploring} leverages a retriever to select demonstrations for image captioning tasks. Additionally, \cite{peng2023icd} propose an ICD-LM to generate demonstrations for a given query, with each demonstration being represented by a token. The effectiveness of this method is evidenced by its performance on image captioning and vision question answering tasks. 
Despite the increasing significance of vision and language generation models in real-world applications, research on RetICL for VL models remains relatively underexplored.

% Designing an efficient and effective fine-tuning strategy is a challenge~\cite{lewis2020retrieval}.  

\section{Conclusion}

This survey concentrates on few-shots In-Context Learning (ICL) using retrieved examples for large language models, a key aspect of Retrieval-Augmented Generation (RAG). We outline various retrieval strategies, diverse retrieval models, retrieval pools, techniques for training demonstration retrievers, and applications. Based on the comprehensive understanding of current trends, we suggest several promising future paths for enhancing the efficacy and functionality of this approach.

\section*{Acknowledge}
We would like to thank Siva Reddy, Suzanna Sia, Andrew Drozdov and Yang Xu for recommending additional references for the initial draft. 
Special thanks are extended to Andrew Drozdov for suggesting the direction of future work on fine-tuning for few-shot learning, and to Xu Yang for suggesting RetICL for VL models.

\bibliography{man}
\bibliographystyle{acl_natbib}

\end{document}